\documentclass[11pt]{article} 
\usepackage{colacl}

\author{Sabine Buchholz \and Jorn Veenstra \and Walter Daelemans\\
ILK, Computational Linguistics, Tilburg University\\
PO box 90153, 5000 LE Tilburg, The Netherlands\\
{\tt[buchholz,veenstra,daelemans]@kub.nl} 
}

\title{Cascaded Grammatical Relation Assignment}

\begin{document} 

\maketitle

\begin{abstract}
In this paper we discuss cascaded Memory-Based grammatical relations
assignment. In the first stages of the cascade, we find chunks of
several types (NP,VP,ADJP,ADVP,PP) and label them with their adverbial
function (e.g.\ local, temporal). In the last stage, we assign grammatical
relations to pairs of chunks. We studied the effect of adding several
levels to this cascaded classifier and we found that even the less
performing chunkers enhanced the performance of the relation finder.
\end{abstract}

\bibliographystyle{acl}

\section{Introduction}
\label{introduction}

When dealing with large amounts of text, finding structure in
sentences is often a useful preprocessing step.  Traditionally, full
parsing is used to find structure in sentences.  However, full parsing
is a complex task and often provides us with more information then we
need.
For many tasks detecting only shallow structures in a sentence
in a fast and reliable way is to be preferred over full parsing. For
example, in information retrieval it can be enough to find only simple
NPs and VPs in a sentence, for information extraction we might also
want to find relations between constituents as for example the subject
and object of a verb.

In this paper we discuss some Memory-Based (MB) shallow parsing
techniques to find labeled chunks and grammatical relations in a
sentence.  Several MB modules have been developed in previous work,
such as: a POS tagger~\cite{Daelemans+96b}, a
chunker~\cite{Veenstra98,TjongKimSang+99} and a grammatical relation
(GR) assigner~\cite{Buchholz98}.  The questions we will answer in this
paper are: Can we reuse these modules in a cascade of classifiers?
What is the effect of cascading?  Will errors at a lower level
percolate to higher modules?

Recently, many people have looked at cascaded and/or shallow parsing and
GR assignment. 
\newcite{Abney91} is one of the first who proposed to
split up parsing into several cascades. He suggests to first find the
chunks and then the dependecies between these chunks.
\newcite{Grefenstette96} describes a cascade of finite-state
transducers, which first finds noun and verb groups, then their heads,
and finally syntactic functions. \newcite{Brants+98} describe a
partially automated annotation tool which constructs a complete parse
of a sentence by recursively adding levels to the tree.
\cite{Collins97,Ratnaparkhi97b} use cascaded processing for full
parsing with good results. \newcite{Argamon+98} applied Memory-Based
Sequence Learning (MBSL) to NP chunking and subject/object
identification. However, their subject and object finders are
independent of their chunker (i.e.\ not cascaded). 

Drawing from this previous work we will explicitly study the effect of
adding steps to the grammatical relations assignment cascade.
Through experiments with cascading several classifiers, we will show
that even using imperfect classifiers can improve overall performance
of the cascaded classifier. We illustrate this claim on the task of
finding grammatical relations (e.g.\ subject, object, locative) to
verbs in text.  The GR assigner uses several sources of information
step by step such as several types of XP chunks (NP, VP, PP, ADJP and
ADVP), and adverbial functions assigned to these chunks (e.g.\
temporal, local). Since not all of these entities are predicted
reliably, it is the question whether each source leads to an
improvement of the overall GR assignment.

In the rest of this paper we will first briefly describe Memory-Based
Learning in Section~\ref{MBL}.  In Section~\ref{chunking}, we discuss
the chunking classifiers that we later use as steps in the cascade.
Section~\ref{relations} describes the basic GR
classifier. Section~\ref{cascade} presents the architecture and
results of the cascaded GR assignment experiments.  We discuss the
results in Section~\ref{discussion} and conclude with
Section~\ref{Conclusion}.

\section{Memory-Based Learning}
\label{MBL}

Memory-Based Learning (MBL) keeps all training data in memory and only
abstracts at classification time by extrapolating a class from the
most similar item(s) in memory. In recent
work \newcite{Daelemans+99} have shown that for typical natural language
processing tasks, this approach is at an advantage because it also
``remembers'' exceptional, low-frequency cases which are useful to
extrapolate from. Moreover, automatic feature weighting in the
similarity metric of an MB learner makes the approach well-suited for
domains with large numbers of features from heterogeneous sources, as
it embodies a smoothing-by-similarity method when data is sparse
\cite{Zavrel+97}.  We have used the following MBL
algorithms\footnote{For the experiments described in this paper we
have used TiMBL, an MBL software package developed in the
ILK-group~\cite{Daelemans+98a},
TiMBL is available from:
http://ilk.kub.nl/.}:

\begin{description}

\item [IB1]: A variant of the $k$-nearest neighbor ($k$-NN) algorithm. The distance between a test item and each
memory item is defined as the number of features for which they have a
different value (overlap metric).

\item [IB1-IG]: IB1 with information gain (an information-theoretic
notion measuring the reduction of uncertainty about the class to be
predicted when knowing the value of a feature) to weight the cost of a
feature value mismatch during comparison.

\item [IGTree]: In this variant, a decision tree is created
with features as tests, and ordered according to the information gain of the 
features, as a heuristic approximation of the computationally more
expensive IB1 variants.

\end{description}
For more references and information about these algorithms we refer to
\cite{Daelemans+98a,Daelemans+99}.
For other memory-based approaches to parsing, see \cite{Bod92} and \cite{Sekine98}.

\section{Methods and Results}
\label{experiments}

In this section we describe the stages of the cascade. The very first
stage consists of a Memory-Based Part-of-Speech Tagger (MBT) for which
we refer to \cite{Daelemans+96b}. The next three stages involve determining
boundaries and labels of chunks.  Chunks are non-recursive,
non-overlapping constituent parts of sentences (see \cite{Abney91}).
First, we simultaneously
chunk sentences into: NP-, VP-, Prep-, ADJP- and
APVP-chunks. As these chunks are non-overlapping, no words can belong to more than one chunk, and thus no conflicts can arise.
Prep-chunks are the prepositional part of PPs, thus
excluding the nominal part.  Then we join a Prep-chunk and one --- or
more coordinated --- NP-chunks into a PP-chunk. Finally, we assign
adverbial function (ADVFUNC) labels (e.g.\ locative or temporal) to
all chunks.

In the last stage of the cascade, we label several types of
grammatical relations between pairs of words in the sentence.

The data for all our experiments was extracted from the Penn Treebank
II Wall Street Journal (WSJ) corpus \cite{Marcus+93}. For
all experiments, we used sections 00--19 as training material and
20--24 as test material. See Section \ref{discussion} for results on other train/test set splittings.

For evaluation of our results we use the precision and recall measures.
Precision is the percentage of predicted
chunks/relations that are actually correct, recall is the percentage
of correct chunks/relations that are actually found.
For convenient comparisons of only one value, we also list the
$F_{\beta=1}$ value \cite{vanRijsbergen79}:
$\frac{(\beta^2+1).prec.rec}{\beta^2 . prec + rec}$, with $\beta=1$

\subsection{Chunking}
\label{chunking}

In the first experiment described in this section, the task is to segment the sentence
into chunks and to assign labels to these chunks. This process of chunking and labeling
is carried out by assigning a tag to each word in a sentence
left-to-right. \newcite{Ramshaw+95} first assigned a chunk tag to each
word in the sentence: {\bf I} for inside a chunk, {\bf O} for outside
a chunk, and {\bf B} for inside a chunk, but the preceding word is in
another chunk. As we want to find more than one kind of chunk, we have to further differentiate the IOB tags as to which kind of chunk (NP, VP, Prep, ADJP or ADVP) the word is in. With the extended
IOB tag set at hand we can tag the sentence:
\begin{tt}

But/CC [NP~the/DT dollar/NN~NP] [ADVP~later/RB~ADVP]
[VP~rebounded/VBD~VP] ,/, [VP~finishing/VBG~VP] [ADJP~slightly/RB
higher/RBR~ADJP] [Prep~against/IN~Prep] [NP~the/DT yen/NNS~NP]
[ADJP~although/IN~ADJP] [ADJP~slightly/RB lower/JJR~ADJP]
[Prep~against/IN~Prep] [NP~the/DT mark/NN~NP] ./.

\end{tt}
as:
\begin{tt}

But/CC$_{O}$ the/DT$_{I-NP}$ dollar/NN$_{I-NP}$ later/RB$_{I-ADVP}$
rebounded/VBD$_{I-VP}$ ,/,$_{O}$ finishing/VBG$_{I-VP}$
slightly/RB$_{I-ADVP}$ higher/RBR$_{I-ADVP}$ against/IN$_{I-Prep}$
the/DT$_{I-NP}$ yen/NNS$_{I-NP}$ although/IN$_{I-ADJP}$
slightly/RB$_{B-ADJP}$ lower/JJR$_{I-ADJP}$ against/IN$_{I-Prep}$
the/DT$_{I-NP}$ mark/NN$_{I-NP}$ ./.$_{O}$

\end{tt}

After having found Prep-, NP- and other chunks, we collapse Preps and
NPs to PPs in a second step. While the GR assigner finds relations
between VPs and other chunks (cf.\ Section~\ref{relations}), the PP
chunker finds relations between prepositions and NPs \footnote{PPs
containing anything else than NPs (e.g. {\em without bringing his
wife}) are not searched for.} in a way similar to GR assignment (see
Section~\ref{relations}).  In the last chunking/labeling step, we assign
adverbial functions to chunks. The classes are the adverbial function
labels from the treebank: LOC (locative), TMP (temporal), DIR
(directional), PRP (purpose and reason), MNR (manner), EXT (extension)
or ``-'' for none of the former.  Table~\ref{ChunkingResults} gives an
overview of the results of the chunking-labeling experiments, using the following algorithms, determined by validation on the train set: IB1-IG for XP-chunking and IGTree for PP-chunking and ADVFUNCs assignment.

\begin{table}[h]
\begin{center}
\begin{tabular}{|l|l|l|l|l|l|l|}
\hline type 	& precision 	& recall 	& $F_{\beta=1}$\\ 
\hline 
\hline
NPchunks		& 92.5		& 92.2		& 92.3\\ 
VPchunks		& 91.9 		& 91.7 		& 91.8 \\
ADJPchunks	& 68.4 		& 65.0 		& 66.7	\\
ADVPchunks	& 78.0		& 77.9		& 77.9 \\
Prepchunks		& 95.5		& 96.7 		& 96.1	\\
\hline
PPchunks		& 91.9 		& 92.2  	& 92.0 \\
\hline
ADVFUNCs	& 78.0 		& 69.5 	  	& 73.5 \\
\hline
\end{tabular}
\end{center}

\caption{Results of chunking--labeling experiments. NP-,VP-, ADJP-, ADVP- and Prep-chunks are found simultaneously, but for convenience, precision and recall values are given separately for each type of chunk.}
\label{ChunkingResults}
\end{table}

\subsection{Grammatical Relation Assignment}
\label{relations}

In grammatical relation assignment we assign a GR to pairs of
 words in a sentence. In our experiments, one of these
 words is always a verb, since this yields the most
 important GRs. The other word is the head of the phrase which is annotated with this grammatical relation in the treebank. A preposition is the head of a PP, a noun of an NP and so on.
Defining relations to hold between heads means that the algorithm can, for example, find a subject relation between a noun and a verb without necessarily having to make decisions about the precise boundaries of the subject NP.

\begin{figure*}[ht]
{\em Not\/}/RB {\em surprisingly\/}/RB ,/, {\em
Peter\/}/NNP {\em Miller\/}/NNP ,/, {\em who\/}/WP
{\em organized\/}/VBD {\em the\/}/DT {\em conference\/}/NN
 {\em in\/}/IN {\em New\/}/NNP {\em
York\/}/NNP ,/, {\em does\/}/VBZ {\em not\/}/RB {\em
want\/}/VB {\em to\/}/TO {\em come\/}/VB {\em
to\/}/IN {\em Paris\/}/NNP {\em
without\/}/IN {\em bringing\/}/VBG {\em his\/}/PRP\$
{\em wife\/}/NN .
\caption{\label{pos_sent} An example sentence annotated with POS.}
\end{figure*}
\setlength{\tabcolsep}{0.1cm}
\begin{table*}[ht]
\begin{tabular}{rrr|rr|rr|rr|rr|rr|r}
 & & & \multicolumn{2}{|c|}{Verb} & \multicolumn{2}{|c|}{Context -2} &
 \multicolumn{2}{|c|}{Context -1} & \multicolumn{2}{|c|}{Focus} &
 \multicolumn{2}{|c|}{Context +1} & Class \\ \hline 
& & & word & pos & word & pos & & word & pos & & word & pos & \\ \hline 
1 & 2 & 3 & 4 & 5 & 6 & 7 & 8 & 9 & 10 & 11 & 12 & 13 & \\ \hline
-7 & 0 & 2 & organized & vbd & - & - & - & - & not & rb & surprisingly & rb & - \\
-6 & 0 & 2 & organized & vbd & - & - & not & rb & surprisingly & rb & , & , &  - \\
-4 & 0 & 1 & organized & vbd & surprisingly & rb & , & , & Peter & nnp & Miller & nnp &  - \\
-3 & 0 & 1 & organized & vbd & , & , & Peter & nnp & Miller & nnp & , & , &  - \\
-1 & 0 & 0 & organized & vbd & Miller & nnp & , & , & who & wp & organized & vbd & np-sbj \\
\end{tabular}
\caption{\label{pos_inst} The first five instances for the sentence in Figure \ref{pos_sent}. Features 1--3 are the Features for distance and intervening VPs and commas. Features 4 and 5 show the verb and its POS. Features 6--7, 8--9 and 12--13 describe the context words, Features 10--11 the focus word. Empty contexts are indicated by the value ``-'' for all features.}
\end{table*}
Suppose we had the POS-tagged sentence shown in Figure \ref{pos_sent}
and we wanted the algorithm to decide whether, and if so how, {\em
Miller} (henceforth: the focus) is related to the first verb {\em
organized}. We then construct an {\em instance} for this pair of words
by extracting a set of feature values from the sentence. The instance
contains information about the {\bf verb} and the {\bf focus}: a
feature for the word form and a feature for the POS of both. It also
has similar features for the local {\bf context} of the
focus. Experiments on the training data suggest an optimal context
width of two elements to the left and one to the right. In the present
case, elements are words or punctuation signs. In addition to the
lexical and the local context information, we include superficial
information about clause {\bf structure}: The first feature indicates
the distance from the verb to the focus, counted in elements. A
negative distance means that the focus is to the left of the verb. The
second feature contains the number of other verbs between the verb and
the focus. The third feature is the number of intervening commas. The
features were chosen by manual ``feature engineering''.  Table
\ref{pos_inst} shows the complete instance for {\em Miller--organized}
in row 5, together with the other first four instances for the
sentence. The class is mostly ``-'', to indicate that the word does
not have a direct grammatical relation to {\em organized}.  Other
possible classes are those from a list of more than 100 different
labels found in the treebank. These are combinations of a syntactic
category and zero, one or more functions, e.g.\ {\tt NP-SBJ} for
subject, {\tt NP-PRD} for predicative object, {\tt NP} for (in)direct
object\footnote{Direct and indirect object NPs have the same label in
the treebank annotation. They can be differentiated by their
position.}, {\tt PP-LOC} for locative PP adjunct, {\tt PP-LOC-CLR} for
subcategorised locative PP, etcetera.  According to their information
gain values, features are ordered with decreasing importance as
follows: 11, 13, 10, 1, 2, 8, 12, 9, 6 , 4 , 7 , 3 , 5.  Intuitively,
this ordering makes sense. The most important feature is the POS of
the focus, because this determines whether it can have a GR to a verb
at all (punctuation cannot) and what kind of relation is possible. The
POS of the following word is important, because e.g.\ a noun followed
by a noun is probably not the head of an NP and will therefore not
have a direct GR to the verb. The word itself may be important if it
is e.g.\ a preposition, a pronoun or a clearly temporal/local
adverb. Features 1 and 2 give some indication of the complexity of the
structure intervening between the focus and the verb. The more complex
this structure, the lower the probability that the focus and the verb
are related. Context further away is less important than near
context. 

\begin{figure*}[ht]
{\bf [ADVP} {\em Not\/}/RB {\em surprisingly\/}/RB {\bf ADVP]} ,/, {\bf [NP} {\em
Peter\/}/NNP {\em Miller\/}/NNP {\bf NP]} ,/, {\bf [NP} {\em who\/}/WP {\bf NP] [VP}
{\em organized\/}/VBD {\bf VP] [NP} {\em the\/}/DT {\em conference\/}/NN {\bf NP]
 \{PP-LOC [Prep} {\em in\/}/IN {\bf Prep] [NP} {\em New\/}/NNP {\em
York\/}/NNP {\bf NP] PP-LOC\}} ,/, {\bf [VP} {\em does\/}/VBZ {\em not\/}/RB {\em
want\/}/VB {\em to\/}/TO {\em come\/}/VB {\bf VP] \{PP-DIR [Prep} {\em
to\/}/IN {\bf Prep] [NP} {\em Paris\/}/NNP {\bf NP] PP-DIR\} [Prep} {\em
without\/}/IN {\bf Prep] [VP} {\em bringing\/}/VBG {\bf VP] [NP} {\em his\/}/PRP\$
{\em wife\/}/NN {\bf NP]} .
\caption{\label{sent} An example sentence annotated with POS (after the slash), chunks (with square and curly brackets) and adverbial functions (after the dash).}
\end{figure*}

\setlength{\tabcolsep}{0.1cm}
\begin{table*}[ht]
\begin{small}
\begin{tabular}{rrr|rr|rrr|rrr|rrrrr|rrr|r}
\multicolumn{3}{c|}{Struct.} & \multicolumn{2}{|c|}{Verb} & \multicolumn{3}{|c|}{Context -2} &
 \multicolumn{3}{|c|}{Context -1} & \multicolumn{5}{|c|}{Focus} &
 \multicolumn{3}{|c|}{Context +1} & Class \\ \hline 
& & & & & word & pos & cat & word & pos & cat & pr & word & pos & cat & adv & word & pos & cat & \\ \hline 
1 & 2 & 3 & 4 & 5 & 6 & 7 & 8 & 9 & 10 & 11 & 12 & 13 & 14 & 15 & 16 & 17 & 18 & 19 & \\ \hline
 -5 & 0 & 2 & org. & vbd & - & - & - & - & - & - & - & surpris. & rb & advp & - & , & , & - & -\\
 -3 & 0 & 1 & org. & vbd & surpris. & rb & advp & , & , & - & - & Miller & nnp & np & - & , & , & - & -\\
 -1 & 0 & 0 & org. & vbd & Miller & nnp & np & , & , & - & - & who & wp & np & - & org. & vbd & vp & np-sbj\\
 1 & 0 & 0 & org. & vbd & who & wp & np & org. & vbd & vp & - & conf.  & nn & np & - & York & nnp & pp & np\\
 2 & 0 & 0 & org. & vbd & org.  & vbd & vp & conf.  & nn & np & in & York & nnp & pp & loc & , & , & - & -\\
\end{tabular}
\caption{\label{inst} The first five instances for the sentence in
Figure \ref{sent}. Features 1--3 are the features for distance and
intervening VPs and commas. Features 4 and 5 show the verb and its
POS. Features 6--8, 9--11 and 17--19 describe the context
words/chunks, Features 12--16 the focus chunk. Empty contexts are
indicated by the ``-'' for all features.}
\end{small}
\end{table*}

To test the effects of the chunking steps from Section~\ref{chunking}
on this task, we will now construct instances based on more structured
input text, like that in Figure \ref{sent}. This time, the focus is
described by five features instead of two, for the additional
information: which type of chunk it is in, what the preposition is if
it is in a PP chunk, and what the adverbial function is, if any. We
still have a context of two elements left, one right, but elements are
now defined to be either chunks, or words outside any chunk, or
punctuation. Each chunk in the context is represented by its last word
(which is the semantically most important word in most cases), by the
POS of the last word, and by the type of chunk. The distance feature
is adapted to the new definition of element, too, and instead of
counting intervening verbs, we now count intervening VP chunks. Figure
\ref{inst} shows the first five instances for the sentence in Figure
\ref{sent}.  Class value``--'' again means ``the focus is not directly
related to the verb'' (but to some other verb or a non-verbal
element). According to
their information gain values, features are ordered in decreasing
importance as follows: 16, 15, 12, 14, 11, 2, 1, 19, 10, 9, 13, 18, 6,
17, 8, 4, 7, 3, 5.
Comparing this to the earlier feature ordering, we see that most of the new features are very important, thereby justifying their introduction. Relative to the other ``old'' features, the structural features 1 and 2 have gained importance, probably because more structure is available in the input to represent.

In principle, we would have to construct one instance for each
possible pair of a verb and a focus word in the sentence. However, we
restrict instances to those where there is at most one other verb/VP
chunk between the verb and the focus, in case the focus precedes the
verb, and no other verb in case the verb precedes the focus. This
restriction allows, for example, for a relative clause on the subject
(as in our example sentence). In the training data, 97.9\% of the
related pairs fulfill this condition (when counting VP
chunks). Experiments on the training data showed that increasing the
admitted number of intervening VP chunks slightly increases recall, at
the cost of precision.  Having constructed all instances from the test
data and from a training set with the same level of partial structure,
we first train the IGTree algorithm, and then let it classify the test
instances. Then, for each test instance that was classified with a
grammatical relation, we check whether the same verb-focus-pair
appears with the same relation in the GR list extracted directly from
the treebank. This gives us the precision of the classifier. Checking
the treebank list versus the classified list yields recall.

\subsection{Cascaded Experiments}
\label{cascade}

We have already seen from the example that the level of structure in
the input text can influence the composition of the instances.  We are
interested in the effects of different sorts of partial structure in
the input data on the classification performance of the final
classifier.

Therefore, we ran a series of experiments. The classification task was
always that of finding grammatical relations to verbs and performance
was always measured by precision and recall on those relations (the test set contained 45825 relations).
The
amount of structure in the input data varied. Table \ref{table_result}
shows the results of the experiments. In the first experiment, only
POS tagged input is used. Then, NP chunks are added. Other sorts of
chunks are inserted at each subsequent step. Finally, the adverbial
function labels are added. We can see that the more structure we add,
the better precision and recall of the grammatical relations get:
precision increases from 60.7\% to 74.8\%, recall from 41.3\% to
67.9\%. This in spite of the fact that the added information is not
always correct, because it was predicted for the test material on the
basis of the training material by the classifiers described in Section
\ref{chunking}. As we have seen in Table \ref{ChunkingResults},
especially ADJP and ADVP chunks and adverbial function labels did not
have very high precision and recall.

\begin{table*}
\begin{center}
\begin{tabular}{|p{4cm}|r|r|r||r|r|r||r|r|r|r|}
\hline
 & & & & \multicolumn{3}{|c||}{All} & Subj. & Obj. & Loc. & Temp. \\ \hline
Structure in input &
\# Feat. & \# Inst. & $\Delta$ & Prec & Rec & $F_{\beta=1}$ & $F_{\beta=1}$ & $F_{\beta=1}$ & $F_{\beta=1}$ & $F_{\beta=1}$ \\ \hline
words and POS only & 
13 & 350091 & 6.1 & 60.7 & 41.3 & {\bf 49.1} & 52.8 & 49.4 & 34.0 & 38.4 \\ \hline
+NP chunks & 
17 & 227995 & 4.2 & 65.9 & 55.7 & 60.4 & 64.1 & 75.6 & 37.9 & 42.1 \\ \hline
+VP chunks & 
17 & 186364 & 4.5 & 72.1 & 62.9 & 67.2 & 78.6 & 75.6 & 40.8 & 46.8 \\ \hline
+ADVP and ADJP chunks & 
17 & 185005 & 4.4 & 72.1 & 63.0 & 67.3 & 78.8 & 75.8 & 40.4 & 46.5 \\ \hline
+Prep chunks &  
17 & 184455 & 4.4 & 72.5 & 64.3 & 68.2 & 81.2 & 75.7 & 40.4 & 47.1 \\ \hline
+PP chunks & 
18 & 149341 & 3.6 & 73.6 & 65.6 & 69.3 & 81.6 & 80.3 & 40.6 & 48.3 \\ \hline
+ADVFUNCs & 
19 & 149341 & 3.6 & 74.8 & 67.9 & {\bf 71.2} & 81.8 & 81.0 & 46.9 & 63.3 \\ \hline
\end{tabular}

\caption{\label{table_result} Results of grammatical relation assignment with more and more structure in the test data added by earlier modules in the cascade. Columns show the number of features in the instances, the number of instances constructed from the test input, the average distance between the verb and the focus element, precision, recall and $F_{\beta=1}$ over all relations, and $F_{\beta=1}$ over some selected relations.}
\end{center}
\end{table*}

\section{Discussion}
\label{discussion}

There are three ways how two cascaded modules can interact. 
\begin{itemize}
\item
The first module can add information on which the later module can
(partially) base its decisions. This is the case between the adverbial
functions finder and the relations finder. The former adds an extra
informative feature to the instances of the latter (Feature 16 in
Table~\ref{inst}). Cf.\ column two of Table \ref{table_result}.
\item
The first module can restrict the number of decisions to be made by
the second one. This is the case in the combination of the chunking
steps and the relations finder. Without the chunker, the relations
finder would have to decide for every word, whether it is the head of
a constituent that bears a relation to the verb. With the chunker, the
relations finder has to make this decision for fewer words, namely
only for those which are the last word in a chunk resp.\ the
preposition of a PP chunk.  Practically, this reduction of the number
of decisions (which translates into a reduction of instances) as can
be seen in the third column of Table \ref{table_result}.
\item
The first module can reduce the number of elements used for the
instances by counting one chunk as just one context element. We can
see the effect in the feature that indicates the distance in elements
between the focus and the verb. The more chunks are used, the smaller
the average absolute distance (see column four
Table~\ref{table_result}).
\end{itemize}

All three effects interact in the cascade we describe. The PP chunker
reduces the number of decisions for the relations finder (instead
of one instance for the preposition and one for the NP chunk, we get only one
instance for the PP chunk), introduces an extra feature (Feature 12 in Table~\ref{inst}), and changes the context (instead of a
preposition and an NP, context may now be one PP).

As we already noted above, precision and recall are monotonically
increasing when adding more structure. However, we note large
differences, such as NP chunks which increase $F_{\beta=1}$ by more
than 10\%, and VP chunks which add another 6.8\%, whereas ADVPs and
ADJPs yield hardly any improvement. This may partially be explained by
the fact that these chunks are less frequent than the former
two. Preps, on the other hand, while hardly reducing the average
distance or the number of instances, improve $F_{\beta=1}$ by nearly
1\%. PPs yield another 1.1\%. What may come as a surprise is that
adverbial functions again increase $F_{\beta=1}$ by nearly 2\%,
despite the fact that $F_{\beta=1}$ for this ADVFUNC assignment step
was not very high. This result shows that cascaded modules need not be
perfect to be useful.

\begin{table*}
\begin{center}
\begin{tabular}{|p{6cm}|l|l|l|}
\hline
Experiment & \multicolumn{3}{|c|}{All Relations} \\ \hline
           &  Precision & Recall & $F_{\beta=1}$ \\ \hline\hline
PP chunking                                                     & 91.9 & 92.2 & 92.0 \\ \hline
PP on perfect test data                                        & 98.5 & 97.4 & 97.9 \\ \hline\hline
ADVFUNC assigmnent                                             & 78.0 & 69.5 & 73.5 \\ \hline
ADVFUNC on perfect test data                                   & 80.9 & 73.4 & 77.0 \\ \hline\hline
GR with all chunks, without ADVFUNC label                      & 73.6 & 65.6 & 69.3 \\ \hline
GR with all chunks, without ADVFUNC label on perfect test data & 80.8 & 73.9 & 77.2 \\ \hline\hline
GR with all chunks and ADVFUNC label                           & 74.8 & 67.9 & 71.2 \\ \hline
GR with all chunks and ADVFUNC label on perfect test data      & 86.3 & 80.8 & 83.5 \\ \hline
\end{tabular}
\end{center}
\caption{\label{table_perf}Comparison of performance of several modules on realistic input (structurally enriched by previous modules in the cascade) vs.\ on ``perfect'' input (enriched with partial treebank annotation). For PPs, this means perfect POS tags and chunk labels/boundaries, for ADVFUNC additionally perfect PP chunks, for GR assignment also perfect ADVFUNC labels.}
\end{table*}

Up to now, we only looked at the overall results. Table
\ref{table_result} also shows individual $F_{\beta=1}$ values for four
selected common grammatical relations: subject NP, (in)direct
object NP,
locative PP adjunct and temporal PP adjunct. Note that the steps have
different effects on the different relations: Adding NPs increases
$F_{\beta=1}$ by 11.3\% for subjects resp.\ 16.2\% for objects, but
only 3.9\% resp.\ 3.7\% for locatives and temporals.  Adverbial
functions are more important for the two adjuncts (+6.3\% resp.\
+15\%) than for the two complements (+0.2\% resp.\ +0.7\%).

\newcite{Argamon+98} report $F_{\beta=1}$ for subject and object
identification of respectively 86.5\% and 83.0\%, compared to 81.8\%
and 81.0\% in this paper. Note however that \newcite{Argamon+98} do
not identify the head of subjects, subjects in embedded clauses, or
subjects and objects related to the verb only through a trace, which
makes their task easier. For a detailed comparison of the two methods
on the same task see \cite{Daelemans+99a}. That paper also shows that
the chunking method proposed here performs about as well as other
methods, and that the
influence of tagging errors on (NP) chunking is less than 1\%.

To study the effect of the errors in the lower modules other than the
tagger, we used ``perfect'' test data in a last experiment, i.e.\ data
annotated with partial information taken directly from the
treebank. The results are shown in Table \ref{table_perf}. We see that
later modules suffer from errors of earlier modules (as could be
expected): $F_{\beta=1}$ of PP chunking is 92\% but could have been
97.9\% if all previous chunks would have been correct (+5.9\%). For
adverbial functions, the difference is 3.5\%. For grammatical relation
assignment, the last module in the cascade, the difference is, not
surprisingly, the largest: 7.9\% for chunks only, 12.3\% for chunks
and ADVFUNCs. The latter percentage shows what could maximally be
gained by further improving the chunker and ADVFUNCs finder. On
realistic data, a realistic ADVFUNCs finder improves GR assigment by
1.9\%. On perfect data, a perfect ADVFUNCs finder increases
performance by 6.3\%.

\section{Conclusion and Future Research}
\label{Conclusion}

In this paper we studied cascaded grammatical relations assignment. We
showed that even the use of imperfect modules improves the overall
result of the cascade.

In future research we plan to also train our classifiers on
imperfectly chunked material. This enables the classifier to better
cope with systematic errors in train and test material.  We expect
that especially an improvement of the adverbial function assignment
will lead to better GR assignment.

Finally, since cascading proved effective for GR assignment we intend to
study the effect of cascading different types of XP chunkers on
chunking performance. We might e.g.\ first find ADJP chunks, then use that chunker's output as additional input for the NP chunker, then use the combined output as input to the VP chunker and so on. Other chunker orderings are possible, too. Likewise, it might be better to find different grammatical relations subsequently, instead of simultaneously.


\end{document}